\documentclass[10pt,journal,compsoc]{IEEEtran}

\usepackage{booktabs}
\usepackage{algorithmic}
\usepackage{algorithm}
\usepackage{amsmath}
\usepackage{amstext}
\usepackage{amssymb}
\usepackage{color}
\usepackage{xcolor}
\usepackage{xspace}
\usepackage{bm}
\usepackage{multirow}
\usepackage{array}
\usepackage{pifont}
\usepackage{makecell}

\newcommand{\hb}[1]{{#1}}

\newcommand{\zzx}[1]{{#1}}

\newcommand{\nothing}[1]{}

\long\def\ignorethis#1{}

\ifCLASSOPTIONcompsoc
  \usepackage[nocompress]{cite}
\else
  \usepackage{cite}
\fi

  \usepackage[pdftex]{graphicx}
  \graphicspath{{./figure}}
  \DeclareGraphicsExtensions{.pdf,.jpeg,.png}

\usepackage{bm}
\usepackage{amsmath}
\usepackage{amsfonts,amssymb}
\usepackage{ragged2e}
\usepackage{steinmetz}

\begin{document}

\title{\emph{MonoNeuralFusion}: Online Monocular Neural 3D Reconstruction with Geometric Priors}

\author{Zi-Xin Zou,
        Shi-Sheng Huang,
        Yan-Pei Cao,
        Tai-Jiang Mu, \\
        Ying Shan, 
        Hongbo Fu
\IEEEcompsocitemizethanks{
\IEEEcompsocthanksitem Zi-Xin Zou and Tai-Jiang Mu
are with BNRist, the Department
of Computer Science and Technology, Tsinghua University,
Beijing, China. 
E-mail: zouzx19@mails.tsinghua.edu.cn, taijiang@tsinghua.edu.cn
\IEEEcompsocthanksitem Shi-Sheng Huang is with the School of Artificial Intelligence, Beijing Normal University, Beijing, China. Email: huangss@bnu.edu.cn.

\IEEEcompsocthanksitem Hongbo Fu is with the School of Creative Media, City University of Hong Kong, Hong Kong, China. Email: hongbofu@cityu.edu.hk.

\IEEEcompsocthanksitem Yan-Pei Cao and Ying Shan are with the ARC Lab, Tencent PCG. Email: caoyanpei@gmail.com, yingsshan@tencent.com.

}

}



\IEEEtitleabstractindextext{%
\begin{abstract}\justifying
High-fidelity 3D scene reconstruction from monocular video\hb{s} continues to be \hb{challenging}, 
especially for \emph{complete} and \emph{fine-grained} geometry reconstruction. 
The previous 3D reconstruction \hb{approaches} with neural implicit representation\hb{s} 
have shown a promising ability for \emph{complete} scene reconstruction, while \hb{their results are often over-smooth and lack enough geometric details.} 
\hb{This paper introduces} 
a novel neural implicit 
scene representation with volume rendering \hb{for high-fidelity online} 
3D scene reconstruction from monocular videos. 
For \emph{fine-grained} reconstruction, our key insight is to incorporate geometric priors into both the neural implicit scene representation and neural volume rendering, thus leading to \hb{an} effective geometry learning mechanism based on volume rendering optimization. 
Benefiting from this, we present \emph{MonoNeuralFusion} to perform the online neural 3D reconstruction from monocular video\hb{s}, by which the 3D scene geometry is efficiently generated and optimized during the on-the-fly 3D monocular scanning. 
\hb{The extensive comparisons with state-of-the-art} 
approaches \hb{show that} our \emph{MonoNeuralFusion} 
consistently generates much better \emph{complete} and \emph{fine-grained} \hb{reconstruction results}, 
both quantitatively and qualitatively. 
\end{abstract}

}

\maketitle

\IEEEdisplaynontitleabstractindextext

\IEEEpeerreviewmaketitle

\begin{figure*}
    \centering
    \includegraphics[width=0.98\linewidth]{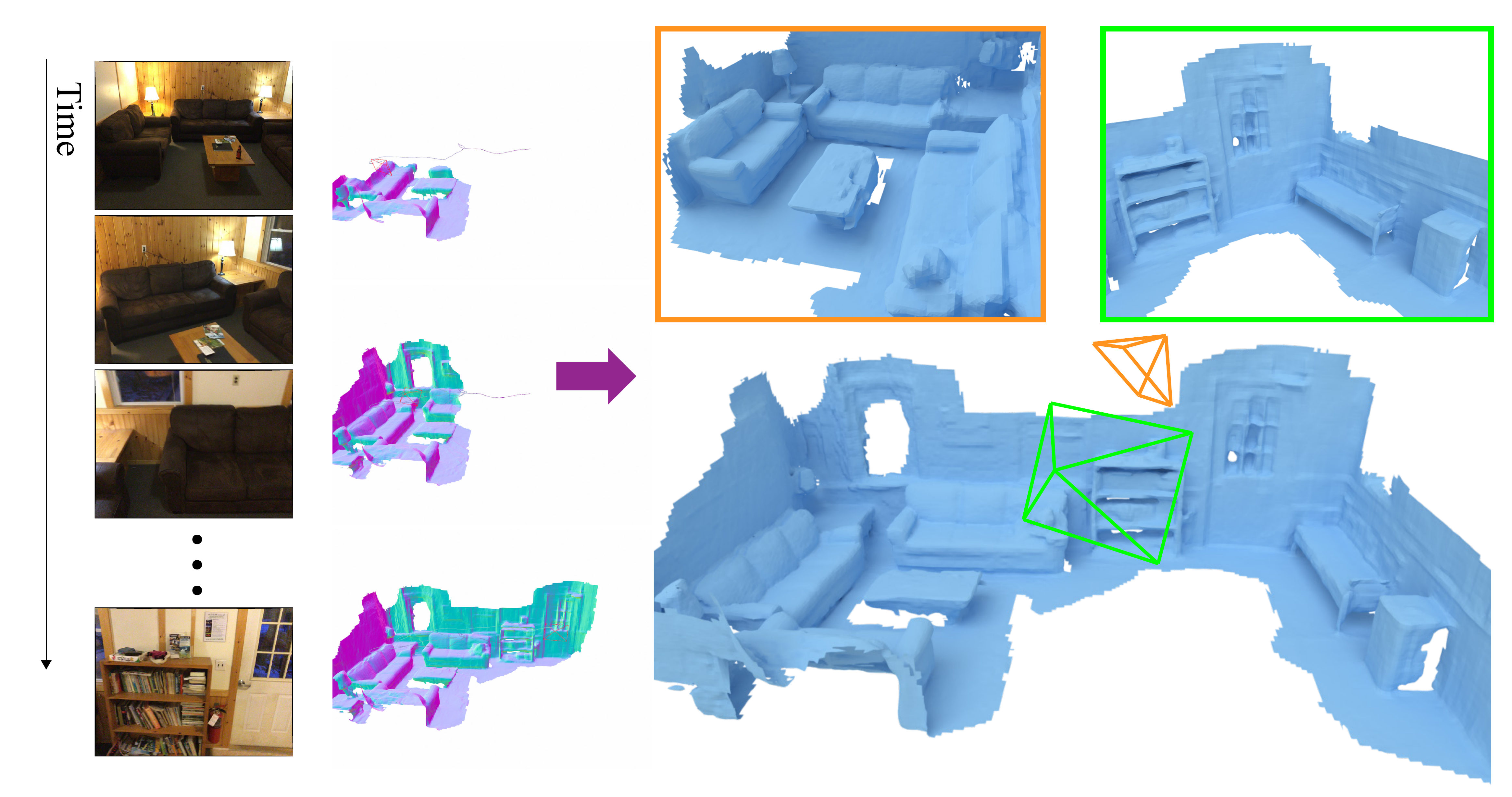}
    \caption{\textbf{Online 3D \hb{scene}  reconstruction from \hb{a} monocular video using our MonoNeuralFusion.} Our MonoNeuralFusion utilizes a neural implicit 
    scene representation with volume rendering and incrementally builds 
    surface reconstruction \hb{from a sequence of posed RGB images by a geometry learning mechanism guided by geometric priors}. 
    \hb{Such a mechanism enables} 
    high-fidelity surface reconstruction with fine geometric details. We illustrate \hb{the} final mesh and \hb{two close-ups} 
    from two views. 
    }
    \label{fig:teaser}
\end{figure*}

\IEEEraisesectionheading{\section{Introduction}\label{sec_intro}}

\IEEEPARstart{O}{nline} reconstruction of 3D indoor scenes from monocular videos continues to be an important research topic in the computer graphics and computer vision communities, \hb{and benefits various} 
applications in virtual/augmented reality, robotics, video games, etc. Although the state-of-the-art visual simultaneous localization and mapping (vSLAM) techniques~\cite{DBLP:journals/trob/QinLS18,DBLP:journals/trob/CamposERMT21} can calculate accurate camera poses from monocular images, it is still challenging for the current monocular 3D reconstruction \hb{solutions} to achieve \emph{complete}, \emph{coherent} and \emph{fine-grained} \hb{reconstruction results}. 

Most depth-based online 3D reconstruction approaches perform 3D volumetric fusion \hb{based on depth maps}, with
\hb{a} surface represented by 
truncated signed distance function (TSDF)~\cite{KinectFusion,BundleFusion,Cao_TOG_2018,9662197,DBLP:journals/tvcg/KahlerPRSTM15} or neural implicit function~\cite{DBLP:conf/cvpr/HuangHS021,Zhu2022CVPR,DBLP:conf/iccv/SucarLOD21}.  
However, one main drawback \hb{of a} 
monocular video input is the \hb{lack} 
of physically reliable depth, making it difficult to apply the current mainstream volumetric fusion techniques to \hb{the} monocular 3D reconstruction task. Although the technique of Mutli-View-Stereo (MVS) or Structure-from-Motion (SfM) can provide coherent depth estimation~\cite{DBLP:journals/tog/ValentinKBWDSVC18,DBLP:conf/eccv/SchonbergerZFP16,schoenberger2016sfm,agarwal2011building,snavely2006photo}, the resulting semi-dense or sparse depth maps often lead to \emph{incomplete} \hb{reconstruction results. In addition, their time-consuming computation makes them} 
not suitable for interactive applications.  With the progress of deep learning, some pioneering works~\cite{tateno2017cnn,zhi2019scenecode,yang2020mobile3drecon} adopt the single-view depth estimation to monocular 3D reconstruction, \hb{and} 
have achieved impressive surface reconstruction results. However, given effective deep learning based monocular depth estimation approaches~\cite{gordon2019depth,DBLP:conf/iccv/GodardAFB19,DBLP:journals/cvm/Li0X22}, it is still challenging to generate consistent depth estimation across different views, making it difficult to build coherent 3D reconstruction for large-scale VA/AR applications. 

The recent work of NeuralRecon~\cite{DBLP:conf/cvpr/SunXCZB21} proposes to reconstruct the 3D geometry with \hb{a} neural network instead of multi-view depth maps, \hb{and} 
has achieved \emph{coherent} 3D surface reconstruction results from monocular videos. However, \hb{their} 
simple average pooling for multi-view 3D volume feature fusion \hb{often leads to} 
a \emph{over-smooth} geometry reconstruction without \hb{enough} 
geometric details. 
Instead, TransformerFusion~\cite{DBLP:conf/nips/BozicPTDN21} introduces spatial-aware feature fusion for better geometric detail reconstruction\hb{. However, its} 
ability is still limited for \emph{fine-grained} geometry reconstruction of certain objects, such as chair legs, monitor stands, etc. The recent success of NeRF~\cite{DBLP:conf/eccv/MildenhallSTBRN20} 
utilizes 
powerful volume rendering with \hb{an} implicit representation, which enables impressive surface reconstruction from multi-view images. Although subsequent works achieve even better fine-grained surface reconstruction with the aid of geometry regularization (NeuS~\cite{DBLP:conf/nips/WangLLTKW21}, MonoSDF~\cite{Yu2022MonoSDF}, NeuRIS~\cite{wang2022neuris}) or Manhattan-world assumption\cite{guo2022manhattan}, the time-consuming geometry learning of implicit scene representations keeps them away from online surface reconstruction for monocular videos.
\hb{An} efficient and effective geometry learning mechanism \hb{remains unexplored for} 
online monocular 3D reconstruction, towards \emph{fine-grained} surface reconstruction with better geometric details.

\hb{Aiming at much better \emph{fine-grained} surface reconstruction quality during the on-the-fly 3D monocular scans}, 
we introduce a novel neural implicit scene representation with volume rendering \hb{for} 
the online monocular 3D reconstruction task. 
Instead of encoding the 3D scene geometry as a single MLP~\cite{DBLP:conf/eccv/MildenhallSTBRN20,DBLP:conf/nips/WangLLTKW21,DBLP:conf/nips/YarivGKL21,DBLP:conf/nips/LiuGLCT20}, we formulate \hb{a} neural implicit scene representation (NISR), which encodes a 3D scene as sparse feature volumes and decodes it as \hb{a} \emph{\hb{continuous}} 
signed distance function \hb{(SDF)}, \hb{instead of a} 
resolution-dependent \hb{SDF} 
as NeuralRecon~\cite{DBLP:conf/cvpr/SunXCZB21}. Based on NISR, we develop a novel volume rendering {approach} and an \emph{efficient} volume rendering optimization, which is especially suitable for 
incremental geometry learning during the online surface reconstruction task. Moreover, 
we further introduce geometric priors (surface normal prior, eikonal regularization prior, and normal map prior) to both the neural implicit scene representation learning and the volume rendering optimization, leading to an \emph{effective} geometric prior guided geometry learning mechanism for high-fidelity surface reconstruction \hb{with} 
geometric details.

Based on such \hb{an} \emph{efficient} and \emph{effective} geometric prior guided neural implicit scene representation with volume rendering, we propose \emph{MonoNeuralFusion} to perform \emph{coherent} 3D reconstruction \hb{from} 
on-the-fly monocular 3D scans, with much better fine-grained surface quality. 
To demonstrate \hb{its} 
effectiveness, 
we have extensively evaluated our approach on various public 3D indoor scan datasets, such as ScanNet~\cite{DBLP:conf/cvpr/DaiCSHFN17}, TUM-RGBD~\cite{DBLP:conf/iros/SturmEEBC12}, and Replica~\cite{replica19arxiv}, \hb{in comparison with state-of-the-art} 
online monocular 3D reconstruction approaches, such as 
NeuralRecon~\cite{DBLP:conf/cvpr/SunXCZB21} and TransformerFusion~\cite{DBLP:conf/nips/BozicPTDN21}. Results show that our approach \hb{achieves} 
better surface reconstruction in quantitative accuracy metrics, with much better fine-detailed geometric details qualitatively, making itself a new state-of-the-art online monocular 3D reconstruction approach. 
We summarise our main contributions \hb{as follows}: 
\begin{itemize}
    \item We introduce the neural implicit scene representation (NISR) with volume rendering, serving as an efficient scene geometry representation for the online geometry learning task.
    \item We propose an effective geometric prior guided geometry learning mechanism, by leveraging the geometric priors to neural implicit scene representation learning and volume rendering optimization, towards high-fidelity surface reconstruction with geometric details.
    \item We introduce 
    MonoNeuralFusion, an online system to incrementally build 
    surface reconstruction\hb{, achieving} 
    much better \emph{fine-grained} reconstruction quality \hb{thanks to} 
    our geometric prior guidance.
\end{itemize}
\section{Related Work}\label{sec_related}
\textbf{Surface Reconstruction from Monocular Images.}
Previous monocular surface reconstruction approaches 
can be mainly divided into 
two types.
The first type is a depth-based approach, which first estimates depth from single-view or multi-view images and then performs the geometry reconstruction by volumetric fusion. For instance, CNN-SLAM~\cite{tateno2017cnn} is probably the first 
\hb{to perform} 
the monocular 3D reconstruction by predicting \hb{a} depth map from \hb{a} 
single view with a CNN-based network and refining \hb{the} depth through \hb{a}  traditional depth filter.
CodeSLAM~\cite{DBLP:conf/cvpr/BloeschCCLD18} and CodeMapping~\cite{matsuki2021codemapping} propose a compact and optimizable code to represent the depth by using a conditional variational auto-encoder(VAE) and jointly optimize {it} 
from multi-view dense bundle adjustment. 
Different from depth estimation from {single-view images}, which completely relies on the learning ability of the {depth prediction} network, 
{depth estimation from} multi-view images could {achieve} \hb{locally} more coherent depth estimation. 
Mobile3DRecon~\cite{yang2020mobile3drecon} uses \hb{a} multi-view semi-global matching method followed by a depth refinement post-processing for robust monocular depth estimation. Recently, learning-based multi-view depth estimation methods benefit from priors with \hb{a} data-driven approach and achieve much better depth estimation quality~\cite{DBLP:conf/3dim/WangS18}.
Bayesian filtering~\cite{DBLP:conf/cvpr/LiuGKNK19}, Gaussian processing~\cite{DBLP:conf/iccv/HouKS19}, and ConvLSTM~\cite{DBLP:conf/cvpr/DuzcekerGVSDP21} are further used to propagate past information to improve the global consistency of \hb{reconstruction results}.
{Some following works combine volumetric convolution~\cite{DBLP:conf/3dim/RichSSH21,choe2021volumefusion} or readily available metadata~\cite{sayed2022simplerecon} with MVSNet, 
producing} \hb{globally more coherent results} \hb{and}
\hb{outperforming the} 
other depth-based methods.

The second type is \hb{a} volume-based method, which directly generates \hb{a} volumetric representation such as \hb{an} occupancy field or SDF field. SurfaceNet~\cite{DBLP:conf/iccv/JiGZLF17} 
proposes to back-project color from two input views to build a color volume and predicts the occupancy probability of the volume grid by using 3D CNNs. Atlas~\cite{murez2020atlas} uses deep image features extracted from 2D CNNs instead of color and extends this method to multi-view images. VoRTX~\cite{DBLP:conf/3dim/StierRSH21} replaces the average-based feature fusion by using \hb{a} transformer architecture~\cite{DBLP:conf/nips/VaswaniSPUJGKP17}. However, all the 
methods mentioned above are performed 
offline. 
To adapt to online applications, NeuralRecon~\cite{DBLP:conf/cvpr/SunXCZB21} and TransformerFusion~\cite{DBLP:conf/nips/BozicPTDN21} perform the incremental feature fusion using the gated recurrent unit (GRU) and transformer, respectively. Compared to \hb{the} depth-based method, \hb{the} volume-based method could achieve globally coherent reconstruction but usually lacks local details. Thus, our 
goal is to improve the level of detail in reconstruction \hb{results}. 
Different from the aforementioned volume-based methods, which focus on improving feature fusion, we focus on the geometric details themselves and leverage the effective geometric prior guided geometry learning mechanism to improve \hb{the} quality of online reconstruction.

\textbf{Neural Implicit Representation.}
Neural implicit representation has shown 
promising results in surface reconstruction in recent years due to its continuous representation and ability \hb{to learn} 
geometric priors from large datasets. DeepSDF~\cite{DBLP:conf/cvpr/ParkFSNL19} and Occupancy Network~\cite{DBLP:conf/cvpr/MeschederONNG19}, for the first time, propose to formulate \hb{an} 
implicit function as a Multi-Layer Perceptron (MLP) with global feature\hb{s} to predict SDF or \hb{an} occupancy value for each query point. 
Some following works divide space into voxels~\cite{DBLP:conf/cvpr/JiangSMHNF20,DBLP:conf/eccv/ChabraLISSLN20,DBLP:conf/eccv/PengNMP020,DBLP:conf/cvpr/HuangHS021,DBLP:journals/corr/abs-2111-12905} or multi-layer voxels~\cite{Zhu2022CVPR,DBLP:conf/cvpr/ChibaneAP20,mueller2022instant} to improve the ability of reconstructing complex geometry with details. 
Some recent online systems, such as DI-Fusion~\cite{DBLP:conf/cvpr/HuangHS021} and NICE-SLAM~\cite{Zhu2022CVPR}, also take advantage 
of neural implicit representation for surface reconstruction from RGB-D sequence\hb{s}. TransformerFusion~\cite{DBLP:conf/nips/BozicPTDN21} is probably the first approach 
based on a neural implicit representation with \hb{an} occupancy field for \hb{the} online 3D monocular reconstruction task.
Unlike TransformerFusion, we perform the geometry learning with the guidance of geometric priors instead, which helps to learn finer surface geometric details.

\textbf{Neural Volume Rendering.}
NeRF~\cite{DBLP:conf/eccv/MildenhallSTBRN20} brings a boom of neural volume rendering in \hb{the} novel view synthesis task, and its variants improve it in terms of speed~\cite{DBLP:conf/iccv/GarbinK0SV21}, representation~\cite{DBLP:conf/nips/LiuGLCT20}, sampling strategy~\cite{DBLP:conf/iccv/BarronMTHMS21}, camera pose~\cite{DBLP:conf/iccv/LinM0L21}, generalization across scenes~\cite{zhang2022nerfusion}, etc. Some works~\cite{DBLP:conf/nips/WangLLTKW21,DBLP:conf/nips/YarivGKL21,DBLP:conf/iccv/OechsleP021} adapt this technology into neural implicit surface\hb{s} to achieve high-fidelity reconstruction from RGB images. \cite{guo2022manhattan} and \cite{Yu2022MonoSDF} additionally use semantic or geometric cues to improve reconstruction quality in indoor scenes. NeuRIS~\cite{wang2022neuris} further improves the results by checking the multi-view consistency to eliminate the effects of \hb{unreliably} 
predicted normals. These methods require hours of optimization and \hb{are} 
time-consuming. When depth sensor\hb{s are} 
available, iMAP~\cite{DBLP:conf/iccv/SucarLOD21} and NICE-SLAM~\cite{Zhu2022CVPR} are two representative methods that apply volume rendering to real-time SLAM system\hb{s}. 
Inspired from these works, our method utilizes 
volume rendering to improve the geometric details of online scene reconstruction by considering 
surface normals. As for as we know, 
no other work is leveraging 
volume rendering on \hb{the} online monocular 3D reconstruction task.
\begin{figure*}[t]
    \includegraphics[scale=0.25]{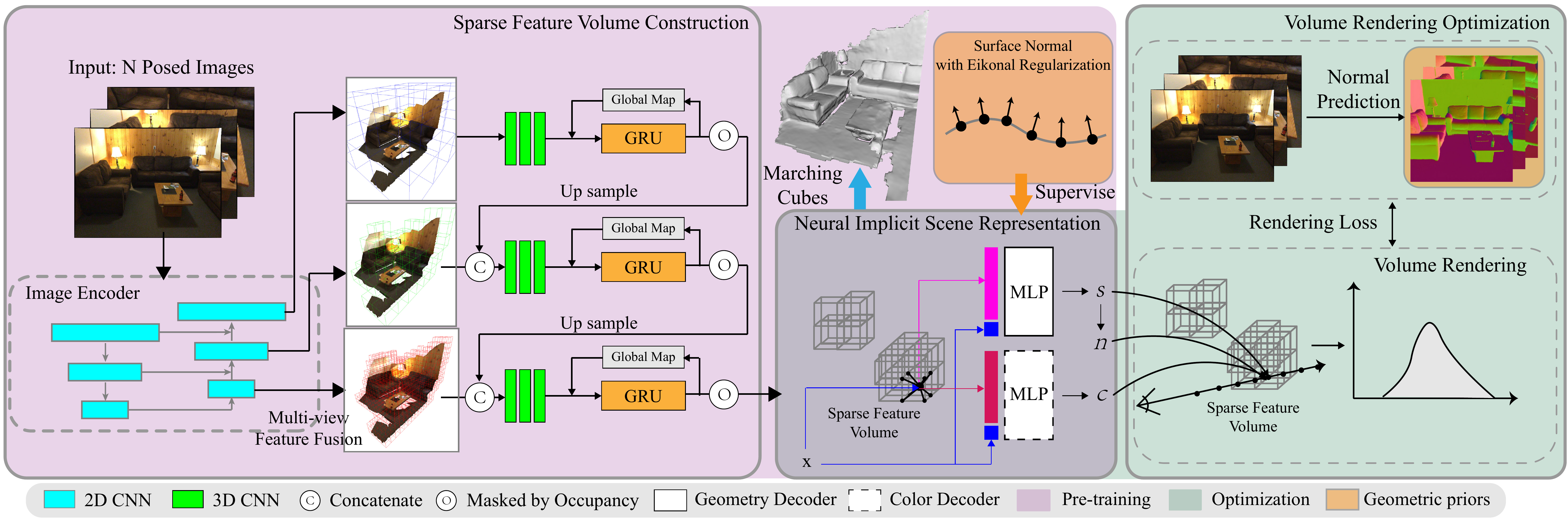}
    \caption{\hb{The pipeline of} \textbf{MonoNeuralFusion.}
    {
    Our approach introduces a flexible and effective neural implicit scene representation (NISR), which encodes 3D scene geometry as a sparse feature volume. Based on this representation, we propose a volume rendering approach suitable for such \hb{a} sparse feature volume. The NISR can be {either} 
    constructed from multi-view images in a coarse-to-fine fashion through \hb{a} pre-trained encoder (purple area) {or} refined by volume rendering optimization (green area). Furthermore, geometric prior guided geometry learning is applied to take full advantage of geometric prior{s} both in pre-training and volume rendering optimization for high-fidelity surface reconstruction.
    }
    }
    \label{fig:overview}
\end{figure*}

\section{MonoNeuralFusion}\label{sec_method}
Given a sequence of $N$ monocular RGB images $\{I_n\}_{n=1}^N$ with \hb{the corresponding} camera poses $\{T_n\}_{n=1}^N$, our goal is to perform 
incremental geometry learning for monocular 3D reconstruction with high quality and fine details.

To this end, we first formulate \hb{a} neural implicit scene representation (NISR) (Sec.~\ref{sec_method:scene_representation}), which encodes 3D scene geometry as \hb{a} sparse feature volume extracted from multi-view images in a coarse-to-fine fashion, and decodes it as \hb{a continuous SDF}.  
Based on this, we further introduce a volume rendering \hb{technique} (Sec.~\ref{sec_method:rendering}), with a hierarchical sampling that is suitable for \hb{the} NISR representation.
For fine-grained surface reconstruction, we develop \hb{a} 
geometry learning mechanism, by leveraging geometric priors both in NISR pre-training and volume rendering optimization (Sec.~\ref{sec_method:training_and_optimization}).
Finally, we provide an online system 
(Sec.~\ref{sec_method:system}) to process incremental surface mapping for coherent reconstruction with fine-grained geometric details.
Fig.~\ref{fig:overview} shows the main components of MonoNeuralFusion, and Fig.~\ref{fig:system} illustrates the pipeline of our online reconstruction system.

\subsection{Neural Implicit Scene Representation}\label{sec_method:scene_representation}
Most of previous online \hb{reconstruction} works represent the 3D geometry as resolution-dependent SDF\hb{s} 
\cite{DBLP:conf/cvpr/SunXCZB21,murez2020atlas} or explicit depth {maps}~\cite{DBLP:conf/iccv/HouKS19,DBLP:conf/cvpr/DuzcekerGVSDP21}.
These two representations have 
disadvantages in the ability of detail expression and \hb{result coherence}, 
compared with neural implicit representation\hb{s}, and \hb{they are not suitable for} 
volume rendering optimization.
Although some methods based on volume rendering optimization achieve impressive and fine-grained surface reconstruction~\cite{DBLP:conf/nips/WangLLTKW21,Yu2022MonoSDF,guo2022manhattan,wang2022neuris}, \hb{they require} 
a \hb{time-consuming} 
optimization from scratch and is hard to extend to \hb{an} online version without a good initial prediction.
Besides, those encoding {an entire 
scene} 
to the parameters of MLP as global feature\hb{s} is difficult to \hb{learn} 
general geometry priors from datasets. 
Based on the above, to both leverage the advantages of stronger expression ability of neural implicit scene representation\hb{s} with volume rendering for fine-grained reconstruction, we propose to formulate a neural implicit scene representation (NISR) for \hb{a} more flexible and effective geometry representation.

Our NISR encodes \hb{an} 
entire 3D scene as \hb{a} 
sparse feature volume, each voxel \hb{containing} 
a continuous \hb{SDF} 
decoded by a 
latent vector fused from image {features} 
of multi-view images.
Note that compared with the continuous occupancy field in \cite{DBLP:conf/nips/BozicPTDN21}, the continuous SDF 
has \hb{a} stronger ability \hb{for detail} 
expression and is reasonable to supervise on \hb{the} gradient domain. Besides, the sparse structure would lead to more \hb{efficency and effectivness} 
for higher resolution.
Finally, the surface mesh can be extracted using Marching Cubes~\cite{DBLP:conf/siggraph/LorensenC87} at an arbitrary resolution. 
One benefit of our NISR is \hb{the enabled} 
efficient geometry learning based on volume rendering optimization in the space of latent vectors, which is suitable for the online 3D reconstruction task. Besides, we further pre-train our NISR with the guidance of geometric priors, leading to more effective feature latent vector extraction for 
fine-grained surface reconstruction.

\textbf{Sparse Feature Volume Construction.}
We adopt the sparse feature volume data structure from NeuralRecon~\cite{DBLP:conf/cvpr/SunXCZB21} to organize the 3D scene's geometry content, where the entire 3D space is divided into a set of sparse voxels\hb{. Each} 
voxel defines a neural \hb{SDF} 
decoded by a feature latent vector {trilinearly} 
interpolated by \hb{the} features from \hb{its} eight corners.
Specifically, we gradually construct the sparse feature volumes $F_\theta^{l}, l\in\{c,m,f\}$($c,m,f$ denotes \hb{the} coarse, middle and fine level\hb{s, respectively}) in a coarse-to-fine fashion, with $F_\theta^{l}$ \hb{being} 
feature latent vectors 
fused by the 2D CNN-based image encoder from the input multi-view images (Fig.~\ref{fig:overview}).
To further improve the overall quality, we adopt a spatial-aware feature fusion based on a transformer module~\cite{stier2021vortx} instead of channel-wise average feature fusion used in NeuralRecon\hb{. The transformer-based fusion} 
helps to build a more expressive feature latent vector{s} 
$F_\theta^{l}$. These feature latent vectors $F_\theta^{l}$ are further merged with \hb{the} previous feature latent vectors using a GRU module after sparse 3D CNN, as global feature latent vectors~\cite{DBLP:conf/cvpr/SunXCZB21}. For brevity, we also notate the final global feature latent vectors as $F_\theta^{l}$ in each sparse feature volume. 

\textbf{Sparse Feature Volume Decoder.}
We decode the feature latent vector $F_\theta^{f}$ in \hb{the} sparse feature volume in \hb{the} fine level as a continuous SDF, which implicitly represents the geometry content. Specifically,
for any query point $\mathbf{x} \in R^3$, we use an MLP-based decoder $f_\theta$ to predict its signed distance $s$ as:
\begin{equation}\label{eq:mlp_decoder}
    s=f_\theta(\mathbf{x}, interp(F_\theta^f,\mathbf{x})),
\end{equation}
where $interp$ means the trilinear interpolation in $F_\theta^f$.
Besides, we additionally decode a radiance field $F_\omega$ for $F_\theta^f$ using an 
MLP-based radiance decoder $f_\omega$:
\begin{equation}
c=f_\omega(\mathbf{x},interp(F_\omega,\mathbf{x})).
\end{equation}

\subsection{Volume Rendering}\label{sec_method:rendering}
In this section, we propose a novel volume rendering \hb{approach to render} 
the geometry content represented by our NISR to any given novel views. \hb{Besides} 
rendering the color map, we also render a normal map, which brings extra geometric regularization for the later volume rendering optimization. Specifically, given a casting ray $\{{\mathbf{p_i}}=\mathbf{o}+d_i\mathbf{v}\}$, where $\mathbf{o}$ is the camera center and $v$ is the view direction, we render both the color map and the normal map along this casting ray by accumulating the measurements of the $N$ sampling points {(see Fig. \ref{fig:overview})} 
as:
\begin{gather}
\hat{C}(\mathbf{r})=\sum_{i=1}^{N}T_i\alpha_i c_i,  \hat{N}(\mathbf{r})=\sum_{i=1}^{N}T_i\alpha_i n_i, 
\end{gather}
where $\alpha_i$ is the discrete opacity value transformed from the SDF value $s_i$ decoded by $f_\theta$ following NeuS~\cite{DBLP:conf/nips/WangLLTKW21}, and $T_i=\prod_{j=1}^{i-1}(1-\alpha_i)$ is the accumulated transmittance. $c_i$ is the color value decoded by $f_\omega$ and $n_i$ is the normal estimation that can be calculated by the automatic derivation of the SDF decoder $f_\theta$.

\begin{figure}[t]
    \includegraphics[scale=0.16]{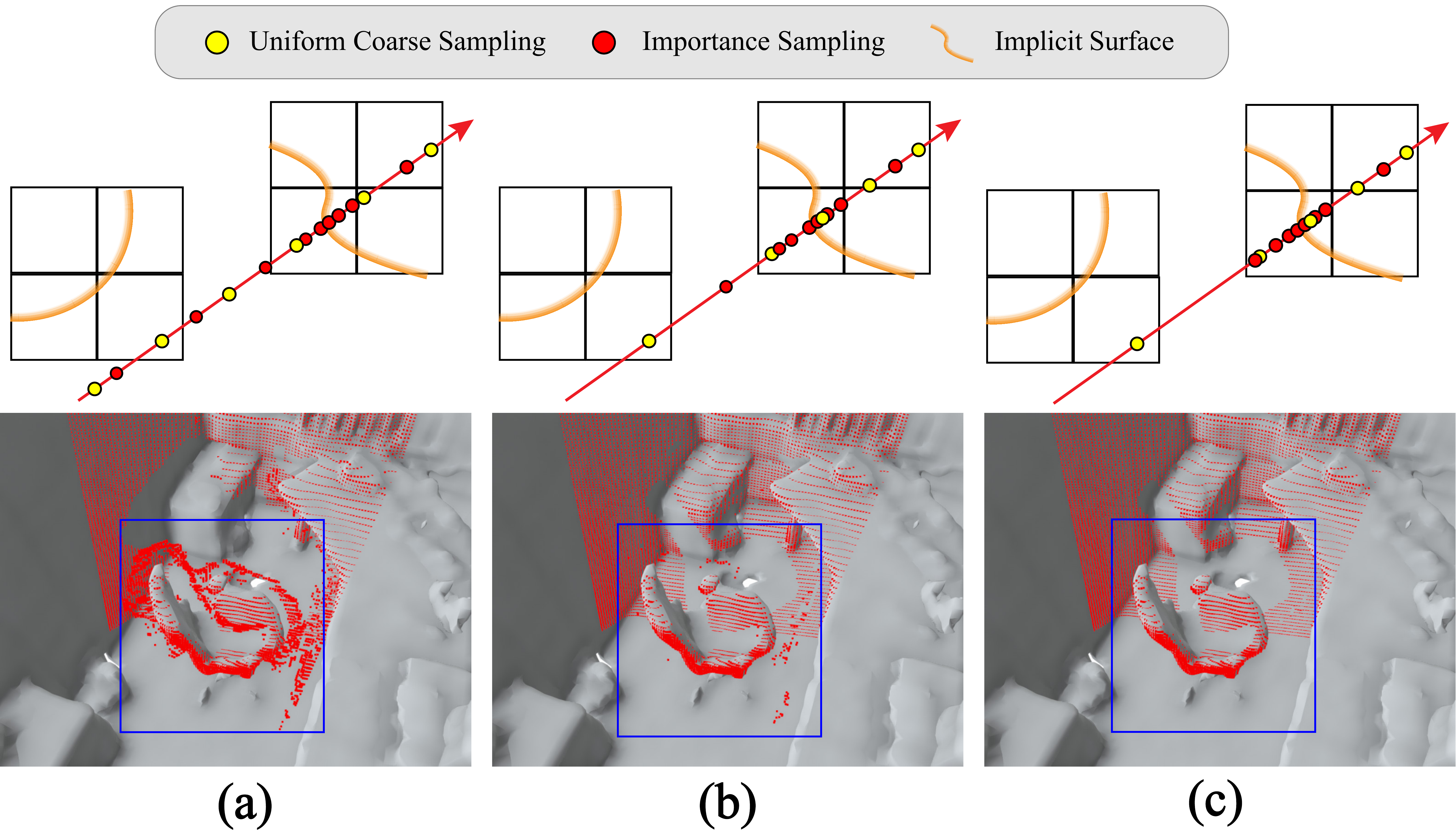}
    \caption{\textbf{Sampling strategies.} {We show the schematic (the first row) and \hb{rendering results} 
    (the second row; the rendered points are colored as red) for three sampling strategies:} (a) Hierarchical sampling strategy without any constraint. (b) Directly applying importance sampling based on coarse sampling strategies inside voxels as NSVF. (c) our hierarchical sampling strategy ensures all sampling points to be inside voxels.
    }
    \label{fig:sampling}
\end{figure}

Since NISR is a sparse feature volume representation, \hb{and} 
is different from the ordinary MLP-based representation like NeRF~\cite{DBLP:conf/eccv/MildenhallSTBRN20} and NeuS~\cite{DBLP:conf/nips/WangLLTKW21}, we propose a hierarchical sampling strategy {(see Fig. \ref{fig:sampling}(c))} 
for a better point sampling \emph{inside} \hb{the} sparse voxels with \hb{the following} two steps: (1) ray-voxel intersection sampling; and (2) inside-voxel sampling.
Specifically, we first perform a ray-voxel intersection sampling to select sparse voxels that \hb{intersect} 
with each casting ray. Then within each intersected \hb{voxel}, 
we perform hierarchical inside-voxel sampling to sample specific points by converting \hb{the} distance {of a sampling point from the camera origin} in \hb{the} world space to \hb{the} sparse space. Note that both uniformly coarse sampling and importance sampling on the top of the coarse probability estimation are processed in \hb{the} sparse space. Given ray-voxel intersection pairs $\mathcal{P}=\{(r,v)\}$, with their distances of ray-voxel out position in the world space $d_{w}^{(r,v)}$ and in \hb{the} sparse space $d_{s}^{(r,v)}$ from the camera origin, we perform distance conversion of any sampling point as follows:
\begin{gather}
    d_s = d_w - (d_w^{(r,v)}-d_s^{(r,v)}), 
\end{gather}
where $v$ is the corresponding intersected voxel id of ray $r$ for \hb{position} sampling. 
In the end, we get a total of $N_c$ coarse sampling points and $N_i$ importance sampling points for each ray rendering.

Applying \hb{the} hierarchical sampling strategy proposed in NeuS without any constraint would result in invalid sampling outside voxels, as \hb{illustrated} 
in Fig.~\ref{fig:sampling}(a).
NSVF, as using a similar scene representation with sparse voxels, yields a self-pruning from dense voxels and samples points guided by sparse voxels which are near the surface as optimization proceeds, thus eliminating the need of hierarchical sampling.
However, sparse voxels would not guide point sampling very well in our method due to its more thickness near \hb{the} underlying surface, and directly applying importance sampling would also lead to invalid sampling outside sparse voxels (Fig.~\ref{fig:sampling}(b)).
Instead, our hierarchical sampling strategy can achieve more accurate point sampling\hb{, i.e.,} 
all samplings locate inside voxels.
{The second row of Fig.~\ref{fig:sampling}(a-c)} illustrates the {rendering results of these three sampling strategies}. We can see that some rendered points (in blue boxes) in {(a) and (b)} are in the air due to \hb{the sampled} 
points out of \hb{the} sparse voxel space, while \hb{the} rendered points of our method are on the surface {(c)}.

\subsection{Geometric Prior Guided Geometry Learning}\label{sec_method:training_and_optimization}
In this subsection, we propose a geometric prior guided geometry learning mechanism, by taking full advantage of the geometric priors both in NISR pre-training and volume rendering optimization for high-fidelity reconstruction.
In the NISR pre-training, we leverage effective regularization from geometric priors, such as normal prior and eikonal prior, to learn the encoder-decoder parameters of our NISR for \hb{a} more detailed surface representation.
In \hb{the} volume rendering optimization, since it does not 
provide ground-truth surface normal\hb{s} sampled from a {ground-truth} mesh 
as in \hb{the} pre-training stage, we leverage the geometric cues from \hb{the} extra normal map prediction from RGB image\hb{s} to enhance the geometric details of reconstruction.

\textbf{NISR Pre-training.} For better feature latent vector extraction, we propose to pre-train our NISR beforehand with \hb{the} geometric prior\hb{s}. Here we formulate the geometric prior 
as the surface normal loss {with eikonal regularization} as:
 \begin{gather}
     \mathcal{L}_{n}=(1-\langle \nabla f_\theta(\mathbf{x}), \hat{n}\rangle) + \lambda_e||\nabla f_\theta(\mathbf{x})^2 - 1||^2,
 \end{gather}
 \hb{where $f_\theta$ is the MLP-based decoder introduced in Equation \ref{eq:mlp_decoder}.} 
We sample points on the surface of \hb{a} ground-truth mesh and compute their normals {$\hat{n}$} for supervision.
Finally, we train our NISR similar to NeuralRecon\cite{DBLP:conf/cvpr/SunXCZB21} but with a geometric prior guided loss function:
\begin{gather}
\mathcal{L}=\mathcal{L}_{o} + \lambda_s\mathcal{L}_{s} + \lambda_n\mathcal{L}_{n} + \lambda_f\mathcal{L}_f, 
\end{gather}
where $\mathcal{L}_o=\sum_{l=1}^{L}\lambda_o^{l}\rm{BCE}(o^{l},\hat{o}^{l})$ denotes the binary cross-entropy (BCE) loss, $\mathcal{L}_{s}=|clamp(s,\tau)-clamp(\hat{s},\tau)|$ represents the clipped L1 loss with $clamp(x,\tau)=\min(\tau, \max(-\tau, x))$, and $\mathcal{L}_f=\frac{1}{M}\sum_{i=1}^{M}||f_i||_2^2$ 
\hb{is} 
used to regularize scene feature vectors $f_i \in F_\theta^l$.

\textbf{Volume Rendering Optimization.} Based on the novel volume rendering from our NISR, we propose to optimize \hb{the} latent vector of \hb{the} sparse feature volume to perform 
accurate geometry learning for high-fidelity surface reconstruction. 
Specifically, we perform the optimization with the following loss function:
\begin{gather}
    \mathcal{L}=\mathcal{L}_{rgb} + \lambda\mathcal{L}_{normal}\label{equ:render_loss}, 
\end{gather}
where $\mathcal{L}_{rgb}=\frac{1}{M}\sum|\hat{C}(\mathbf{r})-C(\mathbf{r})|$ is the color error between \hb{the} render\hb{ed} color $\hat{C}(\mathbf{r})$ and predicted color $C(\mathbf{r})$ commonly used in other reconstruction works{~\cite{DBLP:conf/nips/WangLLTKW21,DBLP:conf/nips/YarivGKL21}}, and $\mathcal{L}_{normal}=\frac{1}{M}\sum\{|\hat{N}(\mathbf{r})-N(\mathbf{r})|+(1-\langle \hat{N}(\mathbf{r}),N(\mathbf{r})\rangle)\}$ describes the L1 distance and \hb{the} cosine angle between \hb{the} render\hb{ed} normal $\hat{N}(\mathbf{r})$ and predicted normal $N(\mathbf{r})$. $M$ is the number of the rendered pixel in a mini-batch.
We use a pretrained out-of-the-box Omnidata model~\cite{DBLP:conf/iccv/EftekharSMZ21} to predict the normal map for each image.

\begin{figure}[ht]
    \centering
    \includegraphics[scale=0.20]{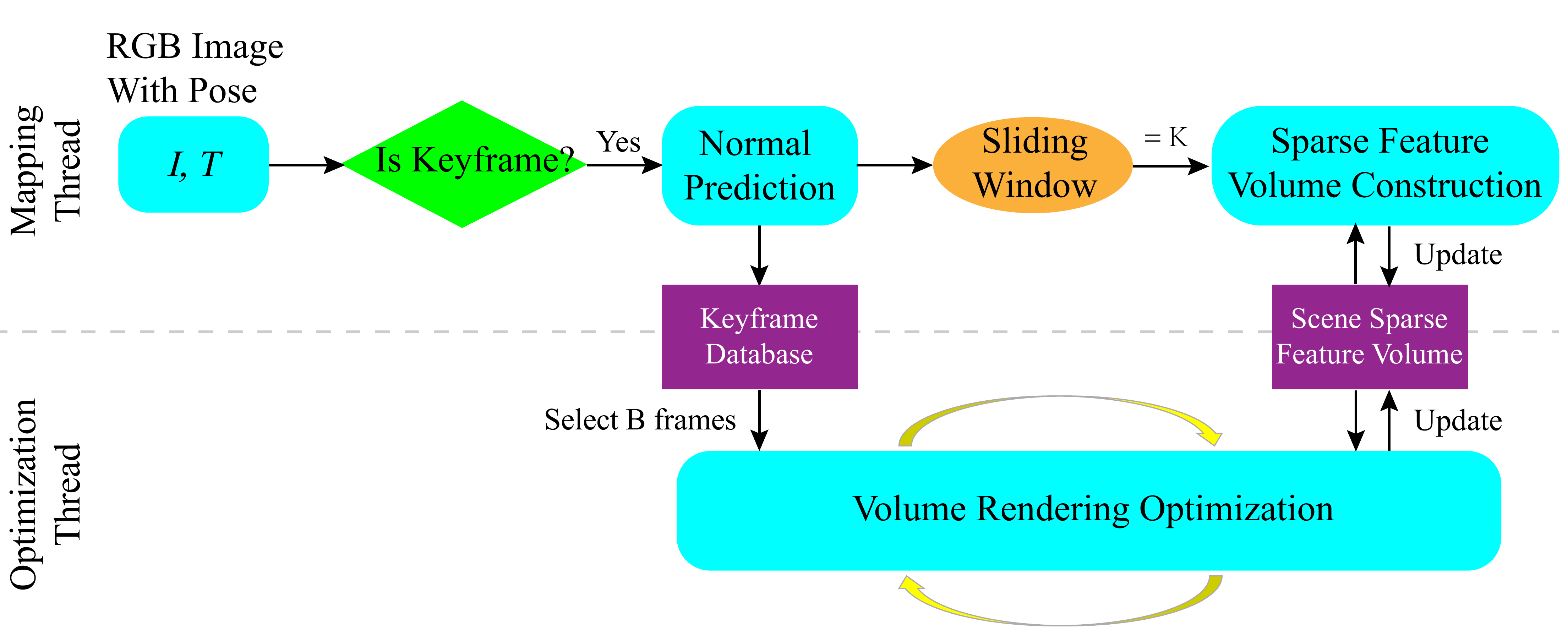}
    \caption{\textbf{Online reconstruction system pipeline.} We run \hb{a} mapping thread and \hb{an} optimization thread in parallel at the back-end to update and refine the feature volume.}
    \label{fig:system}
\end{figure}

\subsection{Online Reconstruction}\label{sec_method:system}
In this subsection, we combine all \hb{the} components introduced above to build an online system to reconstruct the scene geometry.
We build our 
system \hb{similar to a} 
common SLAM system~\cite{DBLP:journals/trob/CamposERMT21}, except \hb{for} the tracking component. In the implementation, We run a mapping thread and an optimization thread in parallel at the back-end\hb{, as illustrated} 
in Fig.~\ref{fig:system}.

\textbf{Mapping Thread.} \hb{The} mapping thread receives the incoming frames from RGB sequences and selects key-frames in sliding windows for efficiency as in NeuralRecon\hb{. Specifically}, 
when the distance between a current frame and the last key-frame is larger than $\tau_d=0.1$ or the rotation angle from the last key-frame is larger than $\tau_a=15^\circ$, a key-frame will be selected. When the size of sliding windows reaches $K=9$, all images and their corresponding camera intrinsic and extrinsic \hb{parameters} will be used to {construct \hb{the} sparse feature volume.}
When \hb{the} feature volume is updated, Marching Cubes \hb{is} 
applied to extract the underlying surface mesh. For every key-frame, it will be applied by a normal prediction and inserted into \hb{a database of} key-frames, 
preparing for the subsequent volume rendering optimization.

\textbf{Optimization Thread.} \hb{The} optimization thread runs loops\hb{. It} 
selects $B$ optimization key-frames in the key-frame database each time and randomly samples $M$ pixels on each of them to build the optimization function as in Equation~\ref{equ:render_loss}. We use Adam optimizer to refine the scene representation parameters for $T$ iterations in a loop.

\textbf{Per-scene Fine-tuning.} To further enhance the results for {a specific} scene, 
we apply an additional fine-tuning for the whole scene. The same as the optimization thread, we optimize \hb{the} scene with sampling $M$ pixels on each of $B$ images for $T$ iterations. Since the reconstruction from \hb{the} online system is already good, a few minutes (less than 9 minutes) fine-tuning is \hb{typically} enough for getting high-fidelity results.
\begin{table*}[ht]
    \centering
    \caption{\textbf{
    \hb{Quantitative comparisons of surface reconstruction} 
    on the ScanNet dataset.} The top block includes \hb{the compared} offline methods while the bottom block includes the compared online methods. The best results are in boldface and the second best are underlined. {Normal consistency of TransformerFusion is missing due to the lack of its publicly released code or mesh results.} 
    }
    \label{tab:3d_metric}
    \begin{tabular}{c |c c c c c c c}
    \toprule[1pt]
        & Acc $\downarrow$ & Comp $\downarrow$ & Chamfer $\downarrow$ & Precision $\uparrow$ & Recall $\uparrow$ & F-score $\uparrow$ & Normal Consistency $\uparrow$ \\
    \midrule
    FastMVSNet & 0.052 & 0.103 & 0.077 & 0.652 & 0.538 & 0.588 & 0.701 \\
    PointMVSNet & 0.048 & 0.115 & 0.082 & 0.677 & 0.536 & 0.595 & 0.695 \\
    Atlas & 0.072 & 0.078 & 0.075 & 0.675 & \textbf{0.609} & 0.638 & 0.819 \\
    \hline
    GPMVS & 0.058 & \textbf{0.078} & \underline{0.068} & 0.621 & 0.543 & 0.578 & 0.715 \\
    DeepVideoMVS & 0.066 & \underline{0.082} & 0.074 & 0.590 & 0.535 & 0.560 & 0.765 \\
    TransformerFusion & 0.055 & 0.083 & 0.069 & 0.728 & 0.600 & \underline{0.655} & - \\
    NeuralRecon & \textbf{0.038} & 0.123 & 0.080 & \underline{0.769} & 0.506 & 0.608 & \underline{0.816} \\
    Ours & \underline{0.039} & 0.094 & \textbf{0.067} & \textbf{0.775} & \underline{0.604} & \textbf{0.677} & \textbf{0.842}
    \end{tabular}
\end{table*}

\begin{table*}[ht]
    \centering
    \caption{\textbf{\hb{Quantitative comparisons of depth estimation} 
    on the ScanNet dataset.} The top block includes the compared offline methods while the bottom block includes the compared online methods. The best results are in boldface and the second best are underlined.}
    \label{tab:2d_metric}
    \begin{tabular}{c|c c c c c c}
    \toprule[1pt]
         & Abs Rel $\downarrow$ & Abs Diff $\downarrow$ & Sq Rel $\downarrow$ & RMSE $\downarrow$ & $\delta < 1.25 \uparrow$ & Comp $\uparrow$ \\
    \midrule
    FastMVSNet & 0.064 & 0.112 & 0.023 & 0.188 & 0.954 & 0.786 \\
    PointMVSNet & \underline{0.057} & \underline{0.098} & \textbf{0.017} & \textbf{0.159} & \textbf{0.965} & 0.683 \\
    Atlas & 0.064 & 0.120 & 0.043 & 0.244 & 0.925 & \underline{0.979} \\
    \hline
    GPMVS & 0.063 & 0.124 & \underline{0.022} & 0.202 & 0.957 & \textbf{1.000} \\
    DeepVideoMVS & 0.061 & 0.128 & \underline{0.022} & 0.204 & \underline{0.962} & \textbf{1.000} \\
    NeuralRecon & 0.066 & 0.099 & 0.038 & 0.197 & 0.932 & 0.891 \\
    Ours & \textbf{0.048} & \textbf{0.079} & 0.024 & \underline{0.164} & 0.951 & 0.921
    \end{tabular}
\end{table*}

\section{Experiments}\label{sec_exp}
In this section, we first give the implementation details of our method and then demonstrate the effectiveness of our method on public datasets, by comparing our approach with \hb{the} other surface reconstruction methods both qualitatively and quantitatively.

\subsection{Implementation Details} 
For the NISR, we use MnasNet~\cite{DBLP:conf/cvpr/TanCPVSHL19} as \hb{a} 2D CNN image encoder to extract multi-level image features and we perform the 3D sparse CNN proposed by~\cite{DBLP:conf/eccv/TangLZLLWH20}. For NISR training, we empirically set the weights of each component in the loss as $\lambda_o^{l}=0.8^{l-1}, \lambda_s=1.0, \lambda_n=1.0, \lambda_e=0.25, \lambda_f=0.001$, and set the clamp threshold $\tau=0.2$.
We train our model using Adam optimizer with an initial learning rate of 0.001. For the volume rendering optimization, we empirically set the weights of normal to be 0.05. 
We select $B=9$ images and sample $M=512$ pixels on each of them as a mini-batch, and sample $N_c=16$ points for coarse {sampling} and $N_i=16$ points {for importance sampling} on each ray and $T=100$ for one loop in our online system. The radiance field $f_\omega$ is initialized randomly from the zero-mean Guassian distribution.
Moreover, we set $N_c=N_i=32$ and optimize the scene in $T=5,000$ iterations for further per-scene fine-tuning.

\subsection{Datasets, Metrics, and Baselines}
\textbf{Datasets.} We select the training subset of ScanNet~\cite{DBLP:conf/cvpr/DaiCSHFN17} as supervision to train our NISR. ScanNet~\cite{DBLP:conf/cvpr/DaiCSHFN17} is a popular real-scan RGB-D dataset, containing 
2.5 million views with ground-truth 3D camera poses and surface reconstruction in more than 1,500 scans. For evaluation, we first evaluate our approach on the testing subset (100 scenes) of ScanNet, but using the monocular RGB sequences only. Additionally, to evaluate \hb{the} generalization ability of our approach, we also perform evaluation on other datasets including TUM RGB-D dataset~\cite{DBLP:conf/iros/SturmEEBC12} (with 10 monocular RGB sequences) and Replica~\cite{replica19arxiv} (with the same 8 monocular sequences by~\cite{DBLP:conf/iccv/SucarLOD21}), using the pre-trained NISR from ScanNet without further fine-tuning on these two datasets.
Since the TUM RGB-D dataset does not 
provide ground-truth 3D surface meshes, we apply TSDF fusion~\cite{DBLP:conf/siggraph/CurlessL96} to generate  3D surface meshes at a resolution of 2cm for the subsequent evaluation.

\textbf{Methods for Comparison.} We compare our method with 
state-of-the-art online monocular reconstruction approaches, including GPMVS~\cite{DBLP:conf/iccv/HouKS19}, DeepVideoMVS~\cite{DBLP:conf/cvpr/DuzcekerGVSDP21},  NeuralRecon~\cite{DBLP:conf/cvpr/SunXCZB21}, and TransformerFusion~\cite{DBLP:conf/nips/BozicPTDN21}, which are 
most relevant to ours. Additionally, to evaluate the final reconstruction quality of our approach, we also compare our approach with some previous offline monocular reconstruction approaches, such as Atlas~\cite{murez2020atlas}, Point-MVSNet~\cite{chen2019point}, and FastMVSNet~\cite{DBLP:conf/cvpr/YuG20}. 
During the comparison, since Atlas, NeuralRecon, TransformerFusion and our approach directly generate a surface mesh as output while others only perform multi-view depth estimation, we fuse the multi-view depth maps into the final surface mesh using TSDF fusion~\cite{DBLP:conf/siggraph/CurlessL96}, thus enabling surface quality comparison for all \hb{the compared} 
approaches.
Besides, we fine-tune those pre-trained models which are not trained on the ScanNet dataset for a fair comparison.

\textbf{Metrics.} To evaluate the surface reconstruction quality, we adopt several popular 3D surface quality metrics, 
including accuracy, completion, chamfer distance, F-score (with both precision and recall)~\cite{murez2020atlas}, and normal consistency~\cite{DBLP:conf/cvpr/MeschederONNG19}. 
For a comprehensive comparison, we additionally 
measure the multi-view depth estimation quality, using the widely used depth map accuracy metrics such as Abs Rel, Abs Diff, Seq Rel, RMSE, $\delta$ and Comp~\cite{DBLP:conf/nips/EigenPF14}. For approaches that do not 
directly generate multi-view depth estimation like ours, we choose to render the  depth maps based on the final surface mesh using pyrender\footnote{https://github.com/mmatl/pyrender}. 

Since TransformerFusion only provides \hb{its} evaluation script without releasing the test code or resulting reconstruction meshes, for a fair comparison we conduct comparison using its evaluation script for 3D metrics on the ScanNet dataset. For the other datasets and 2D metrics evaluation, we use the evaluation script from NeuralRecon, which is different from TransformerFusion's in mesh point sampling.
More details can be found in the supplementary materials.

\begin{figure*}
    \centering
    \includegraphics[scale=0.35]{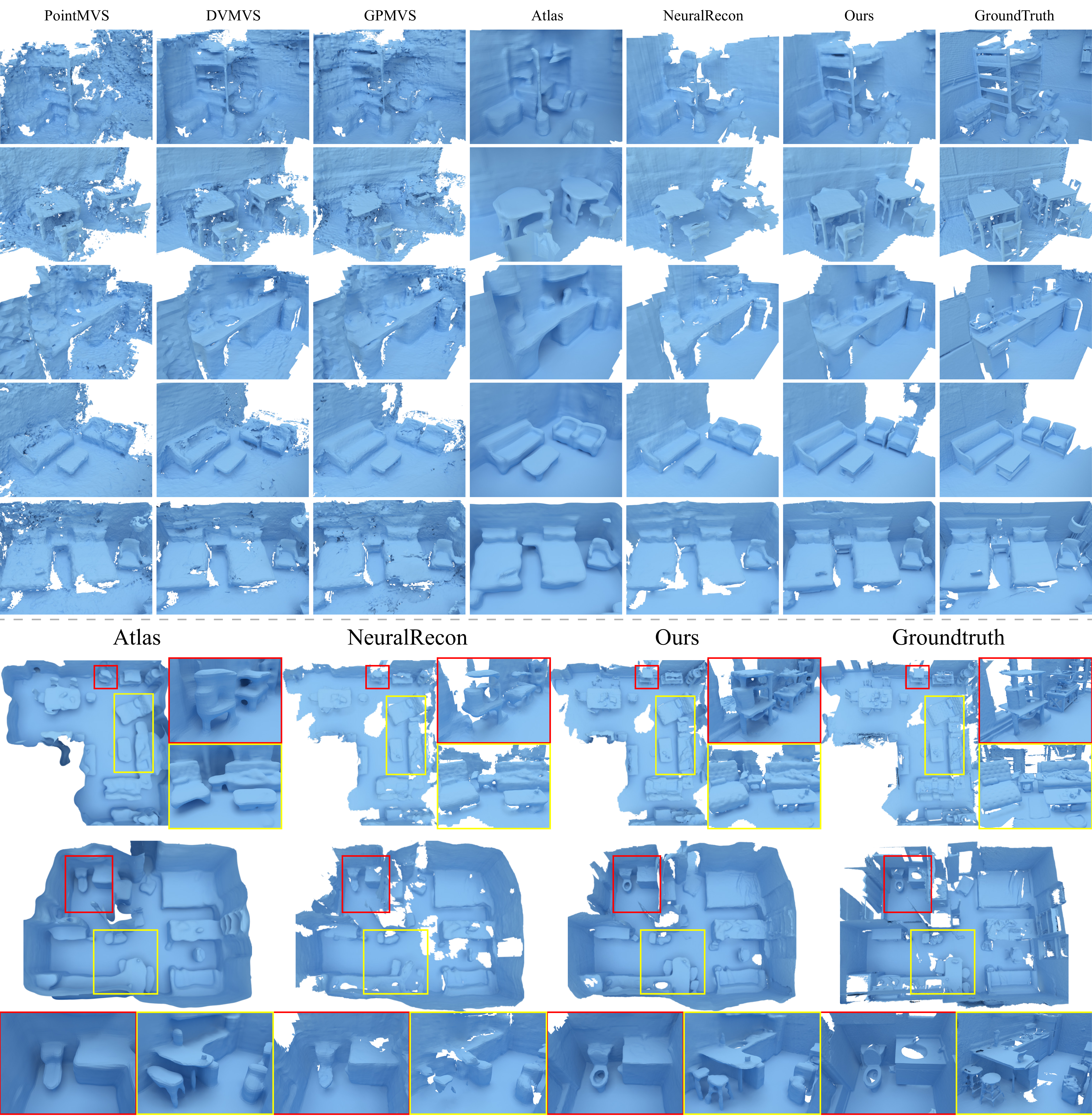}
    \caption{\textbf{Qualitative comparisons on the ScanNet dataset.} We compare our method with three online methods (including GPMVS, DeepVideoMVS, and NeuralRecon) and two offline methods (including Atlas and PointMVS). At the bottom, we \hb{highlight the results of} 
    the best three methods and give a global view for \hb{a} 
    large-scale room with two close views for details.}
    \label{fig:scannet}
\end{figure*}

\subsection{Quantitative and Qualitative Evaluation}

\noindent\textbf{Evaluation on ScanNet.} The quantitative comparison results for surface reconstruction and depth estimation on ScanNet Dataset are shown in Tables~\ref{tab:3d_metric} and~\ref{tab:2d_metric}, respectively. 
Compared to \hb{the} approaches that perform multi-view depth estimation like FastMVSNet, PointMVSNet, GPMVS, and DeepVideoMVS, those 
surface reconstruction approaches {that perform volumetric fusion} including Atlas, NeuralRecon, TransformerFusion and ours can achieve \hb{globally} more 
coherent reconstruction quality with a consistently large improvement in F-score as shown in Table~\ref{tab:3d_metric}.  
It demonstrates that our method achieves the lowest Chamfer Distance {and the highest} F-score and Normal Consistency accuracy values among all \hb{the} 
online and offline methods. {Besides,} almost the same Acc accuracy as NerualRecon (only a very slight increase about 0.001), which 
is much better the other online surface reconstruction approaches.
\hb{As an offline method}, Atlas 
use\hb{s} more views together with dense voxel grids \hb{and} naturally gets higher recall, while our method outperforms NeuralRecon and TransformerFusion \hb{and achieves} 
comparable results on recall to Atlas.
Furthermore, benefiting from our geometric prior guidance in both training and optimization, our method outperforms all the others with a significant improvement in terms of normal consistency, which measures the ability of capturing higher-order information.
The normal consistency of TransformerFusion is missing due to the lack of its publicly released code or reconstruction mesh results.
Since TransformerFusion releases an example mesh of one scene on its Github repository\footnote{https://github.com/AljazBozic/TransformerFusion}, we provide the qualitative and quantitative comparison with it on this scene in the supplementary materials and the comparisons show 
that our method still significantly outperforms TransformerFusion on normal consistency with finer details.
For depth estimation accuracy, our method also significantly outperforms the other methods in terms of Abs Rel and Abs Diff, and achieves 
comparable results in terms of Sq Rel and RMSE, as shown in Table~\ref{tab:2d_metric}.

Fig.~\ref{fig:scannet} shows visual comparison 
results of all the compared online methods and two representative offline methods, including Atlas for a volume-based method and PointMVS for a depth-based method.
It demonstrates that 
the final mesh results of the surface reconstruction approaches are consistently more coherent than multi-view depth estimation methods even in offline configurations. Although NeuralRecon improves details compared with Atlas, it is still far away from fine-grained surface reconstruction. Benefiting from the proposed geometric prior guided NISR with volume rendering, our method not only successfully recovers 
thin parts or small objects, but also achieves sharper geometry (e.g., sofa in the fourth row), significantly improving the detail of reconstruction.

\begin{figure*}[ht]
    \centering
    \includegraphics[scale=0.87]{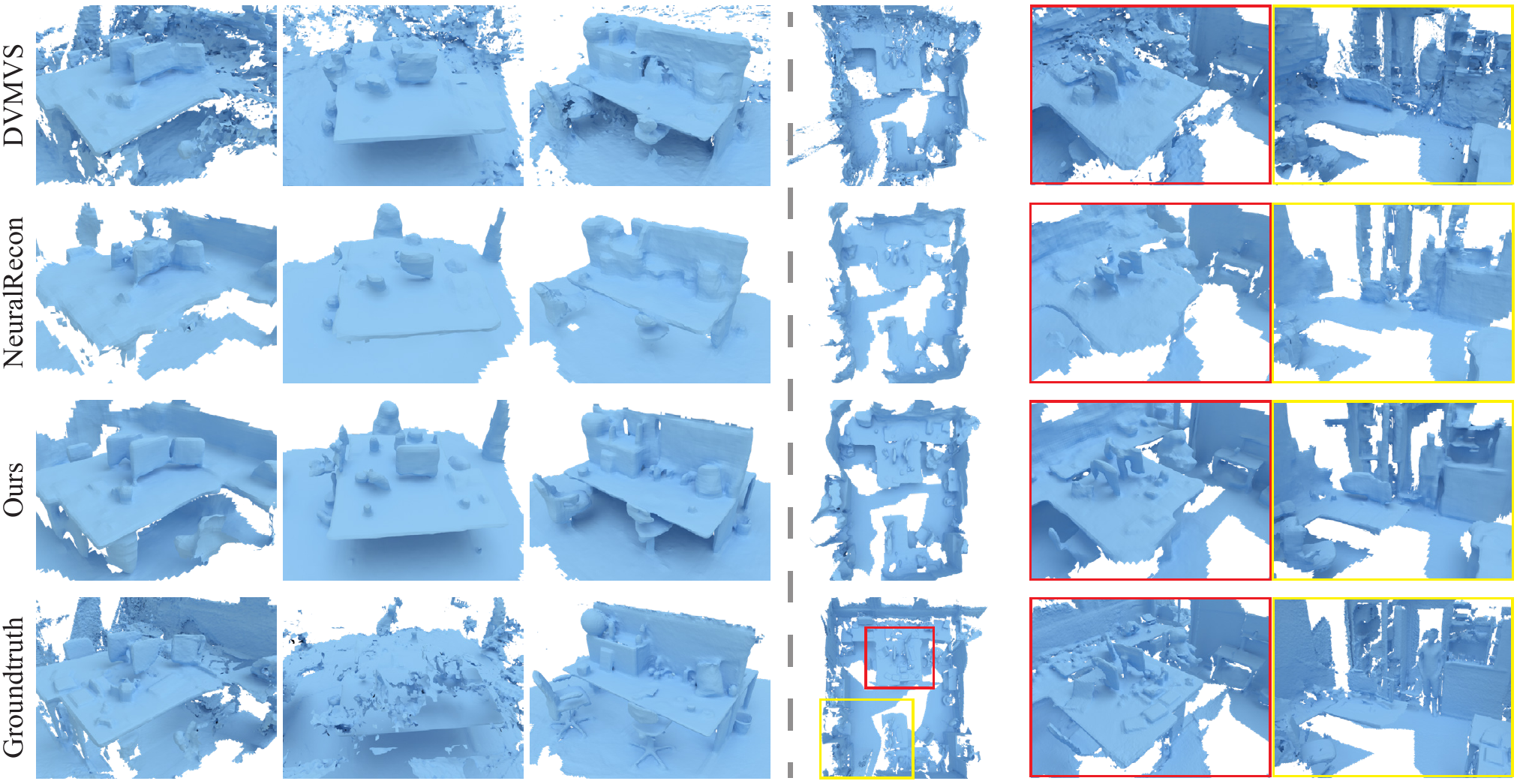}
    \caption{\textbf{Qualitative comparisons on the TUM RGB-D dataset.} Our method still achieves the high-fidelity with more details than all the other methods in a new domain of test dataset. In the right columns, 
    we show the global view with two close-up views (marked by red and yellow boxes).}
    \label{fig:tum-rgbd}
\end{figure*}

\subsection{Generalization on Other Datasets}
\textbf{Evaluation on TUM RGB-D.} We also conduct an evaluation on the TUM RGB-D dataset to show the generalization ability of our model. Table~\ref{tab:tum_rgbd} shows the results on some major metrics compared with the other online methods. 
It can be observed that our method achieves the best performance, 
in \hb{all the} 
surface reconstruction accuracy metrics 
and depth estimation accuracy metrics \hb{except  F-score}.
The main reason of our method achieving a less F-score than GPMVS and DVMVS is the drop on recall (with ours (0.323), GPMVS (0.458), and DVMVS (0.507)). However, our method outperforms them on precision (with ours (0.464), GPMVS (0.343), and DVMVS (0.417)) and also outperforms NeuralRecon both on precison (0.464) and recall (0.258).
For qualitative results, we can see that our method still achieves the high-fidelity results with the richest details (e.g., small objects on the table, chairs and monitors) among the compared methods, 
as shown in Fig.~\ref{fig:tum-rgbd}. 
Note that since the multi-view depth maps estimation would not be always robust for globally coherent surface reconstruction, we only show the visual effect of DeepVideoMVS as the best representative.
This evaluation demonstrates that our geometric prior guided geometry learning for surface reconstruction can be generalized well into a new dataset, despite only pre-trained on the ScanNet dataset.

\begin{table}[t]
    \centering
    \caption{\textbf{Evaluation on \hb{10} selected 
    sequences of the TUM RGB-D dataset.} We conduct a comparison with online methods in 
    major metrics for both 3D (Chamfer distance, F-score, and Normal Consistency) and 2D (Abs Rel, Abs Diff, and first inlier ratio metric).}
    \resizebox{0.5\textwidth}{8mm}{
    \begin{tabular}{c|c c c | c c c}
         & Chamfer $\downarrow$ & F-score $\uparrow$ & N.C. $\uparrow$ & Abs Rel $\downarrow$ & Abs Diff $\downarrow$ & $\delta < 1.25\uparrow$ \\
    \hline
       GPMVS  & 0.201 & \underline{0.387} & 0.649 & 0.079 & 0.202 & 0.915 \\
       DVMVS & 0.152 & \textbf{0.452} & 0.682 & 0.078 & 0.222 & 0.918 \\
       NeuralRecon & 0.134 & 0.325 & 0.788 & 0.090 & 0.140 & 0.928 \\
       Ours & \textbf{0.107} & 0.375 & \textbf{0.806} & \textbf{0.062} & \textbf{0.108} & \textbf{0.959} \\
    \end{tabular}}
    \label{tab:tum_rgbd}
\end{table}

\begin{figure}
    \centering
    \includegraphics[scale=0.58]{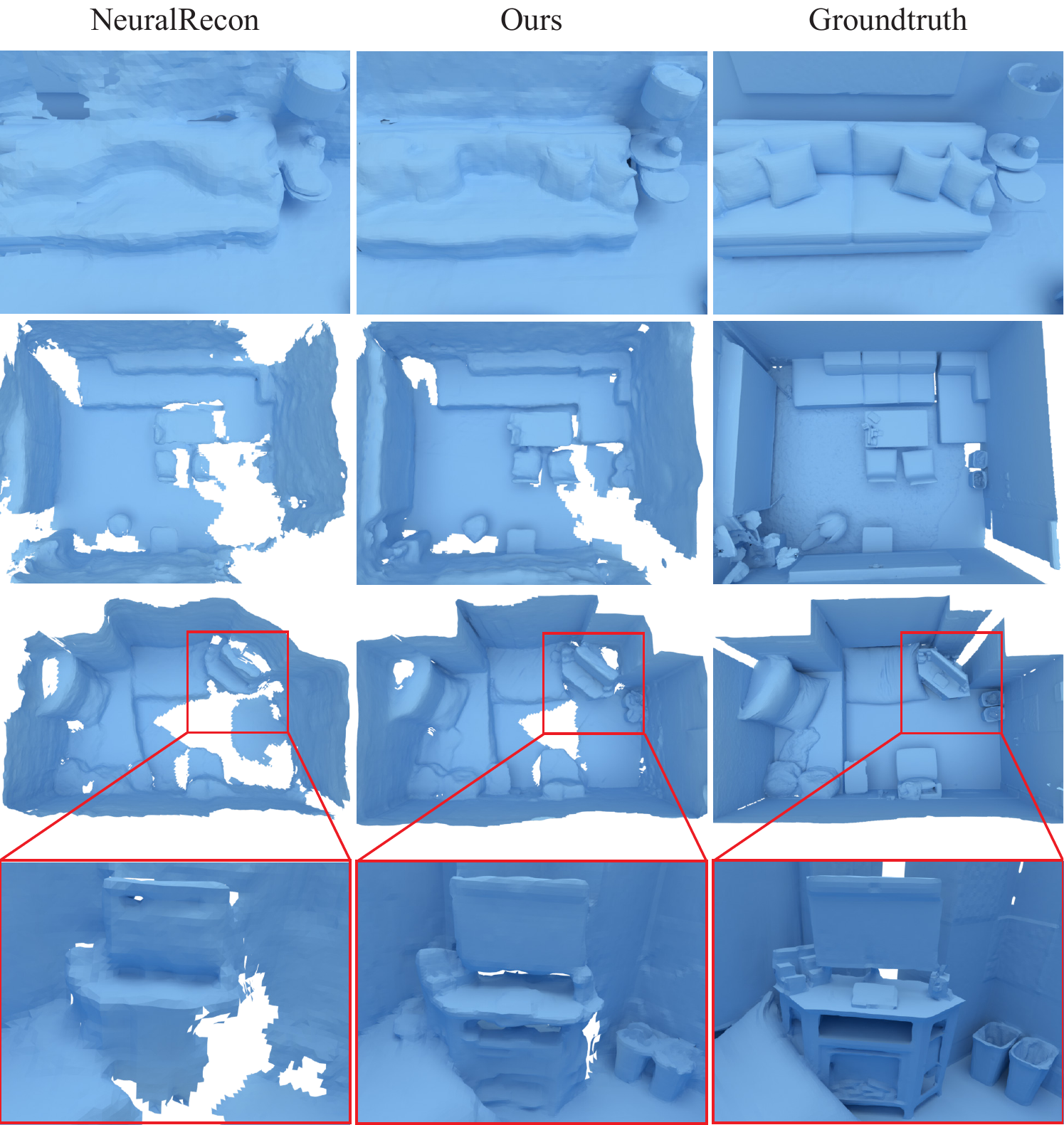}
    \caption{\textbf{Reconstruction results on the Replica dataset.} Compared with NeuralRecon, our method achieve\hb{s} more complete and fine-detailed reconstruction. }
    \label{fig:replica}
\end{figure}

\textbf{Evaluation on Replica.} Similarly, we also evaluate \hb{our results qualitatively in comparison with} 
NeuralRecon on \hb{the} Replica dataset. Fig.~\ref{fig:replica} shows some \hb{comparisons of} visual effects. 
We can see that our method recovers more complete surfaces as well as fine-grained details, such as the pillow on the sofa. Especially, our method successfully reconstructs TV cabinet and garbage can with high-quality while NeuralRecon fails.

\begin{table*}[ht]
    \caption{\textbf{Ablation study on the ScanNet dataset.} 
    We conduct several experiments of different settings as below. \ding{51} and \ding{55} denote using or not using a certain component. $\mathcal{L}_n$ denotes surface normal, ``fusion'' denotes multi-view feature fusion, ``opt'' denotes volume rendering optimization, and ``v.s.'' denotes voxel size (measured in centimeter) of \hb{the} last level of feature volume. Especially, in \hb{the} column of ``fusion'', ``avg'' means average and ``transf'' means transformer. In the column of ``opt'', ``online'' means online volume rendering optimization and ``ft'' means we perform a per-scene fine-tuning afterwards.}
    \centering
    \resizebox{\textwidth}{14mm}{
    \begin{tabular}{c | c c c c |c c c c c c | c c c c c}
    Exp.& $\mathcal{L}_n$ & fusion & opt. & v.s. & Acc. $\downarrow$ & Comp. $\downarrow$ & Prec. $\uparrow$ & Recall $\uparrow$ & F-score $\uparrow$ & N.C. $\uparrow$ & Abs Rel $\downarrow$ & Abs Diff $\downarrow$ & Sq Rel $\downarrow$ & RMSE $\downarrow$ & $\delta < 1.25 \uparrow$  \\
    \midrule
    a & \ding{51} & avg & \ding{55} & 4 & 0.040 & 0.102 & 0.765 & 0.571 & 0.652 & 0.834 & 0.055 & 0.085 & 0.028 & 0.171 & 0.945 \\ 
    b & \ding{55} & transf & \ding{55} & 4 & 0.040 & 0.104 & 0.752 & 0.562 & 0.642 & 0.814 & 0.059 & 0.091 & 0.033 & 0.185 & 0.941\\
    c & \ding{51} & transf & \ding{55} & 4 & 0.039 & 0.095 & 0.775 & 0.599 & 0.674 & 0.841 & 0.051 & 0.082 & 0.026 & 0.170 & 0.949  \\ 
    d & \ding{51} & transf & \ding{55} & 8 & 0.049 & 0.096 & 0.702 & 0.544 & 0.612 & 0.826 & 0.067 & 0.102 & 0.037 & 0.202 & 0.931 \\ 
    e & \ding{51} & transf & \ding{55} & 16 & 0.052 & 0.124 & 0.680 & 0.480 & 0.560 & 0.792 & 0.095 & 0.135 & 0.064 & 0.250 & 0.896 \\ 
    f & \ding{51} & transf & online & 4 & 0.039 & 0.094 & 0.775 & 0.604 & 0.677 & 0.842 & 0.048 & 0.079 & 0.024 & 0.164 & 0.951 \\ 
    g & \ding{51} & transf & ft & 4 & 0.039 & 0.093 & 0.777 & 0.609 & 0.681 & 0.842 & 0.045 & 0.076 & 0.022 & 0.158 & 0.955 \\ 
    
    \end{tabular}
    }
    \label{tab:ablation}
\end{table*}

\subsection{Time Analysis}\label{sec_exp:timing}
We evaluate \hb{the} runtime of our system on a platform with an Nvidia RTX 3090 GPU and Intel Xeon(R) Gold 5218R CPU. Table~\ref{tab:timing} 
provides \hb{the detailed timing} 
of each main component among all the test scenes in ScanNet. 
Our system performs normal prediction for every key-frame in 61.48ms and performs image encoding and feature volume construction for every local fragment (with 9 key-frames) in 34.35ms and 257.13ms, respectively.
For volume rendering optimization, the timing of each iteration (including rendering and backward propagation) is 79.29ms. A final mesh is extracted in 295.79ms when our NISR is updated. Since \hb{the} mapping thread and optimization thread run in parallel, \hb{the} mapping thread can run at up to 2.83 fragments (about 25.5 key-frames) per second.

\begin{figure}[t]
    \centering
    \includegraphics[scale=0.54]{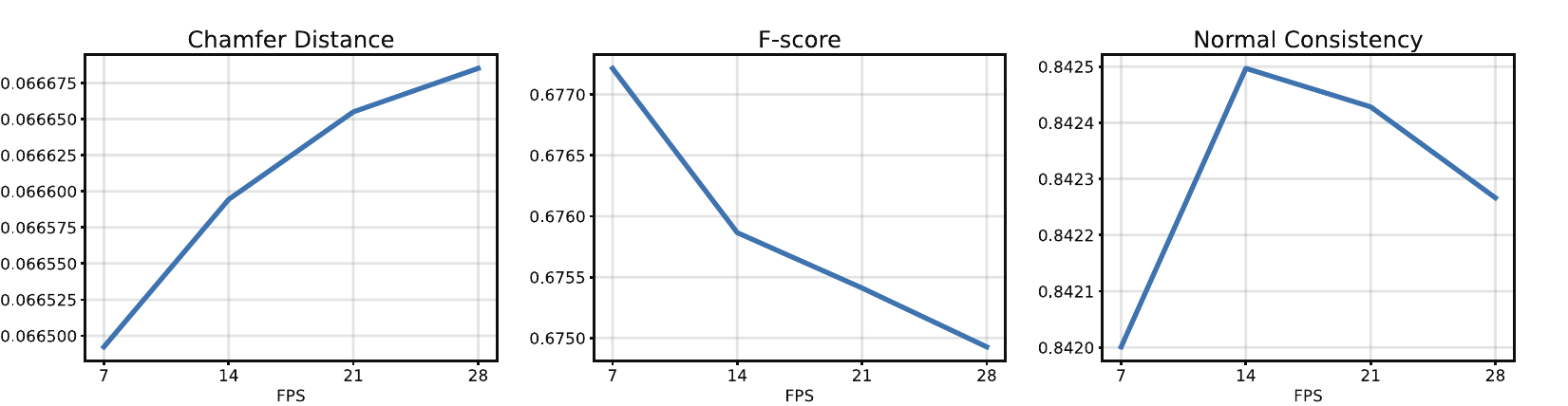}
    \caption{\textbf{Effect of different frame rate.} We demonstrate the curve of quality results on the Chamfer distance, F-score, and normal consistency metrics, respectively, at different interactive frame rate\hb{s}.}
    \label{fig:ab_fps}
\end{figure}

We additionally evaluate the influence of results at different interactive frame rate\hb{s} of input stream and \hb{plot} the curve of quality results on the Chamfer distance, F-score, and normal consistency 
in Fig.~\ref{fig:ab_fps}. We can observe that the more time for optimization with a lower frame rate, the better results it can achieve. Furthermore, it also demonstrates that our method outperforms the NeuralRecon (with the Chamfer distance 0.080, F-score 0.608,
and normal consistency 0.816) and TransformerFusion (with
the Chamfer distance 0.069 and F-score 0.655), regardless of the frame rates.

\begin{table}[t]
    \centering
    \caption{\textbf{Timing of each main component.} ``kf'' is the abbreviation for key-frame, ``frag'' is the abbreviation for local fragment, and ``iter'' is the abbreviation for iterations.}
    \resizebox{0.5\textwidth}{5mm}{
    \begin{tabular}{c|c c c c c}
    Task & \makecell{Normal \\ prediction} & \makecell{Image\\ encoding} & \makecell{Feature volume \\ construction} & Optimization & \makecell{Mesh \\ extraction} \\
    \hline
    Timing (ms) & 61.84/kf & 34.35/frag & 257.13/frag & 79.29/iter & 295.79
    \end{tabular}
    }
    \label{tab:timing}
\end{table}

\subsection{Ablative Analysis}
\begin{figure}
    \centering
    \includegraphics[scale=0.25]{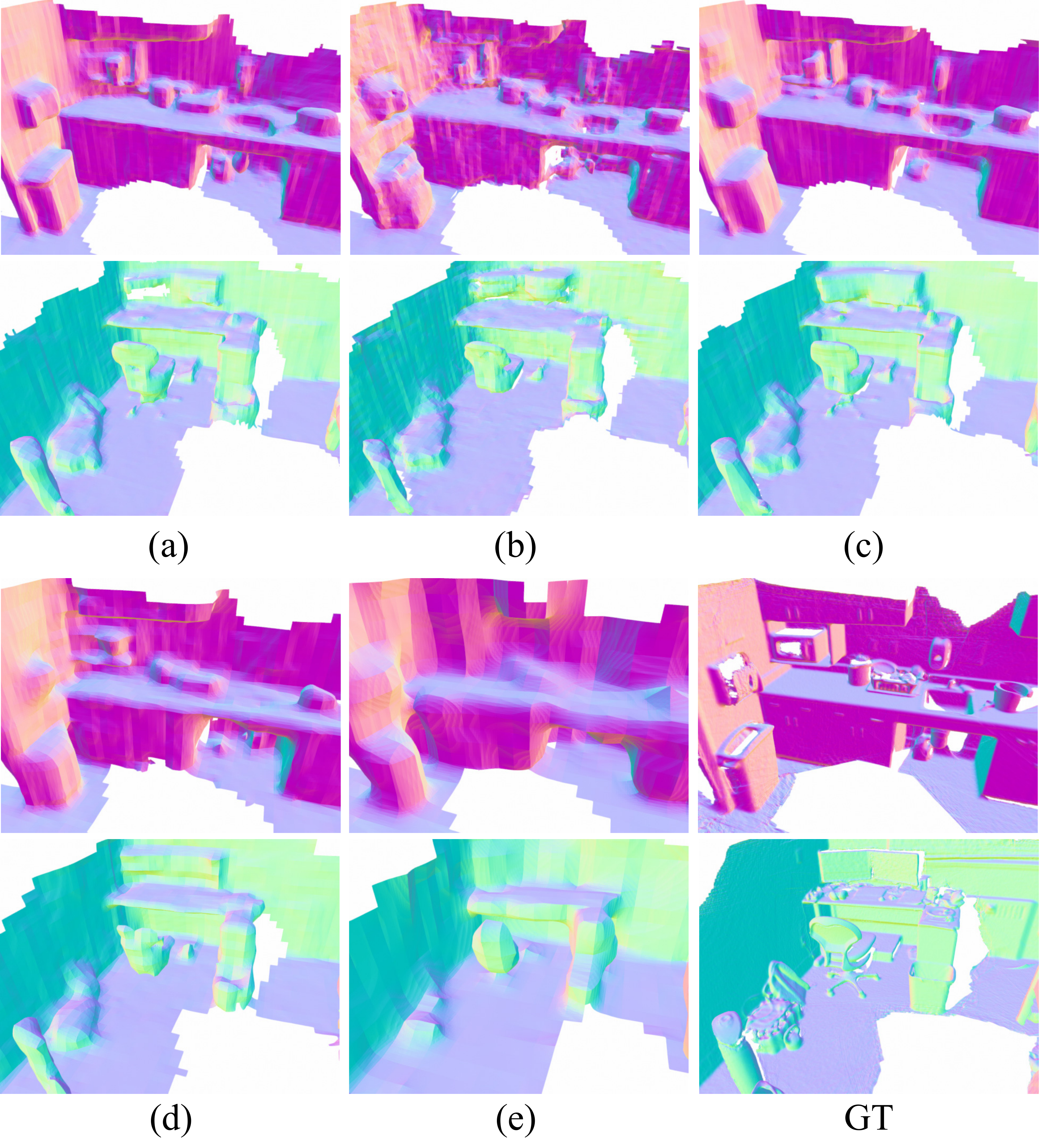}
    \caption{\textbf{Effect\hb{s} of different settings in \hb{the} training stage.} (a)-(e) correspond to (a)-(e) in Table~\ref{tab:ablation}. No surface normal loss would make the surface rough and loss of details, while \hb{the} transformer module is not the core component for \hb{improving the quality of} 
    surface reconstruction. For the voxel size, larger voxel size would lead to the lack of details.}
    \label{fig:ab}
\end{figure}

\begin{figure}
    \centering
    \includegraphics[scale=0.29]{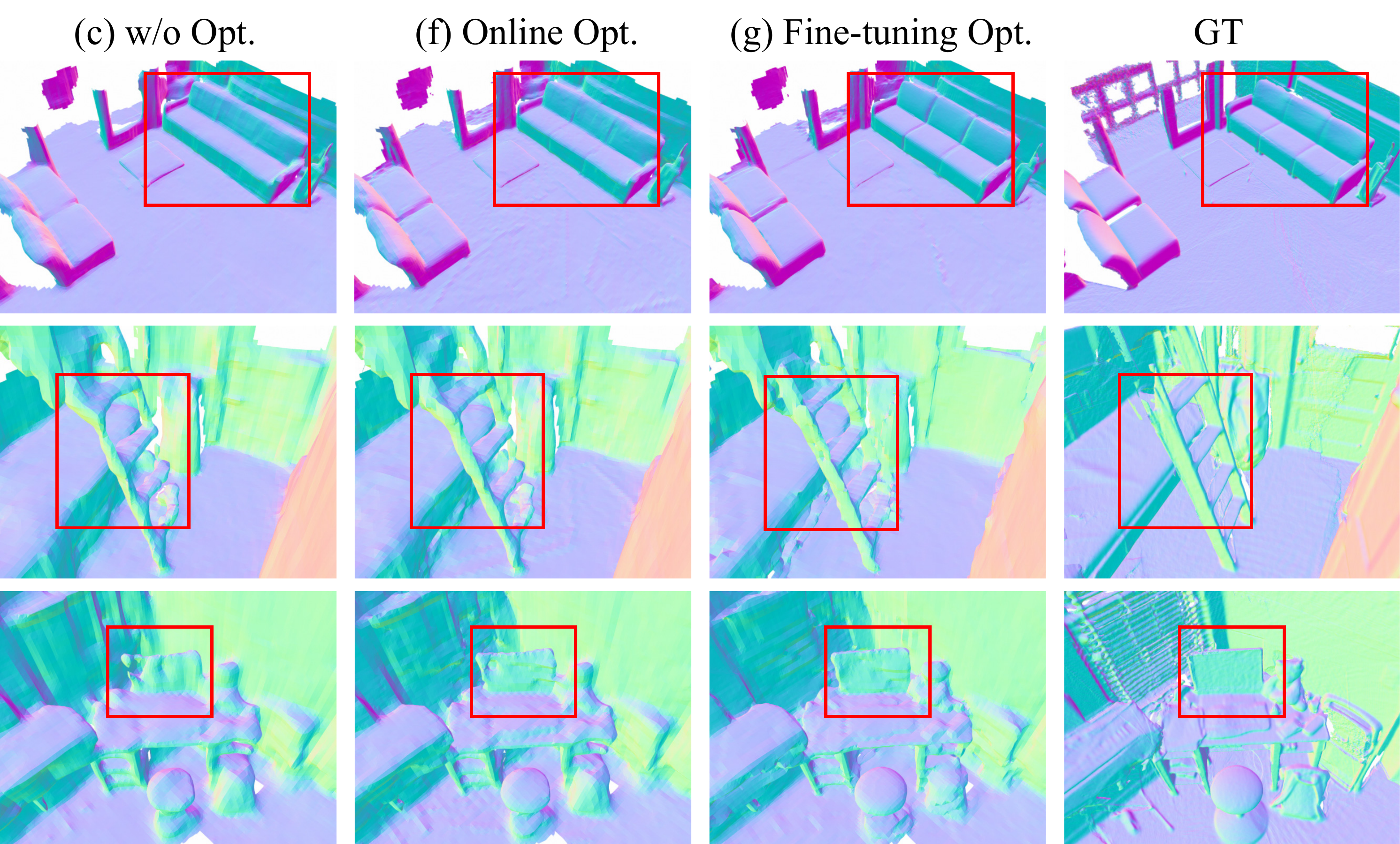}
    \caption{\textbf{Effects of volume rendering optimization to geometric details.} We can see that there are more details of sofa (first row) when \hb{optimizing} the scene, and it also could fill the missing regions of stairs (middle row) and computer display (last row) from model{ing} feed-forward inference. The red boxes highlight the differences of details.}
    \label{fig:ab_opt}
\end{figure}

To demonstrate the effectiveness of each main component of our full system, we conduct an ablation study experiment.
Table~\ref{tab:ablation} lists the quantitative results \hb{under} different settings. 
\zzx{(a)-(g) in the following paragraph denotes the experiment of different settings as illustrated in Table~\ref{tab:ablation} with their quantitative results. Fig.~\ref{fig:ab} and~\ref{fig:ab_opt} show their qualitative results corresponding to (a)-(g) in Table~\ref{tab:ablation}.}

\textbf{Surface Normal Loss.} 
We conduct \hb{an} experiment of not using surface normal (b) and using surface normal (c) to show the effect of \hb{the} geometric prior in \hb{the} training stage.
Table~\ref{tab:ablation} shows 
when removing \hb{the} surface normal loss, F-score and normal consistency (N.C.) of (b) show a significant decrease. 
Fig.~\ref{fig:ab} shows the visual effect without (b) or with (c) this geometric prior. We can observe that the reconstruction surface becomes rough and loses geometric details when removing it.
In conclusion, the surface normal 
as the geometric prior could improve details and regularize the NISR for \hb{smoother surfaces}, 
playing an important role in \hb{the} training stage for high-fidelity reconstruction results.

\textbf{Multi-view Feature Fusion.} We evaluate the effect of different types of multi-view feature fusion with channel-wise average (a) or average from \hb{the} output of transformer blocks (c)
. Although we apply \hb{the} 
transformer blocks in local fragment\hb{s} instead of global region\hb{s} used in \cite{DBLP:conf/3dim/StierRSH21}, Table~\ref{tab:ablation}(c) shows its effectiveness in feature fusion for improving the final result. From Fig.~\ref{fig:ab}, we can see that the improvement of multi-view feature fusion cannot 
directly influence the qualitative results of surface reconstruction, where \hb{the reconstructed} surface \zzx{(see Fig.~\ref{fig:ab}(b))}
is still rough and \hb{lacks} 
details. Thus, surface normal is a more relevant factor for fine-grained quality surface reconstruction, while \hb{the} transformer module is more about improving sparse voxel prediction, leading to a higher recall.

\textbf{Voxel Size.} Although we can obtain a continuous SDF 
from our NISR and the final mesh can be extracted at \hb{an} arbitrary resolution, the voxel size of \hb{the} sparse feature volume also influences the quality of reconstruction. {(c), (d), (e) in} Table~\ref{tab:ablation} and Fig.~\ref{fig:ab}
show the influence of different voxel sizes for the fine level (using 4cm, 8cm and 16cm respectively). It demonstrates that a smaller voxel size will lead to higher quality and more details of reconstruction. Theoretically, further a smaller voxel size can obtain even better results, but it also causes more timing and memory consumption. In our experiment, we find that setting voxel size as 4cm is fine for surface reconstruction in high quality with enough geometric details.

\textbf{Volume Rendering Optimization.} Volume rendering optimization provide a refinement for the sparse feature volume of scene to improve the quality of reconstruction. Compared with (c), using online optimization (f)
would improve the recall (e.g. filling some missing regions of a computer display as shown in Fig.~\ref{fig:ab_opt}) and thereby increase the F-score value in Table~\ref{tab:ablation}. Besides, geometry cues from normal map prediction could add more details with a \hb{slight improvement of} 
normal consistency 
and enhance the visual effects (e.g., adding sofa crevice in Fig.~\ref{fig:ab_opt}). Furthermore, we also provide a per-scene fine-tuning (g) with a whole scene optimization in only a few minutes (less than 9 minutes). This would further slightly improve \hb{the surface reconstruction results both quantitatively and qualitatively}. 

\begin{figure}
    \centering
    \includegraphics[scale=0.175]{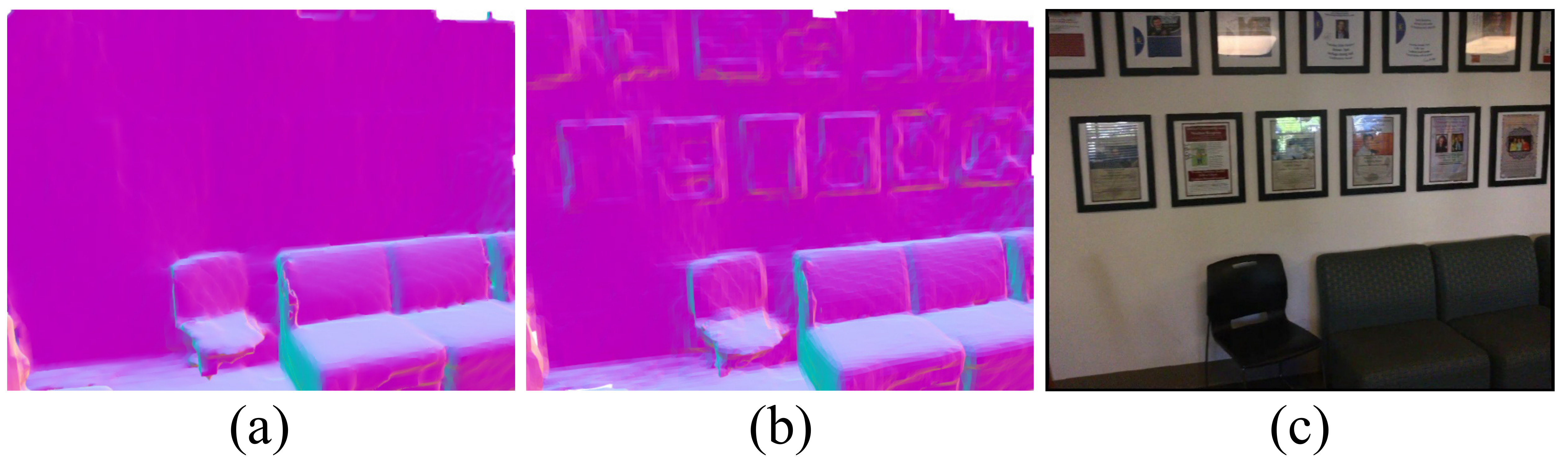}
    \caption{\textbf{Effects of color representation in the optimization step.} (a) optimization without the color loss $\mathcal{L}_{rgb}$. (b) optimization with the color loss $\mathcal{L}_{rgb}$. (c) color image for reference. Optimization with the color representation could obtain richer details.}
    \label{fig:ab_color}
\end{figure}

\textbf{Color Representation.} 
We evaluate the effect of color representation in the optimization step. Although the radiance field cannot 
be optimized well in such a short time, the color representation could provide some details which the normal map does not 
have, further improving the \hb{quality of} surface reconstruction. Fig.~\ref{fig:ab_color} shows that the reconstruction result has more details when using \hb{the} color representation.

\begin{figure}[t]
    \centering
    \includegraphics[scale=0.25]{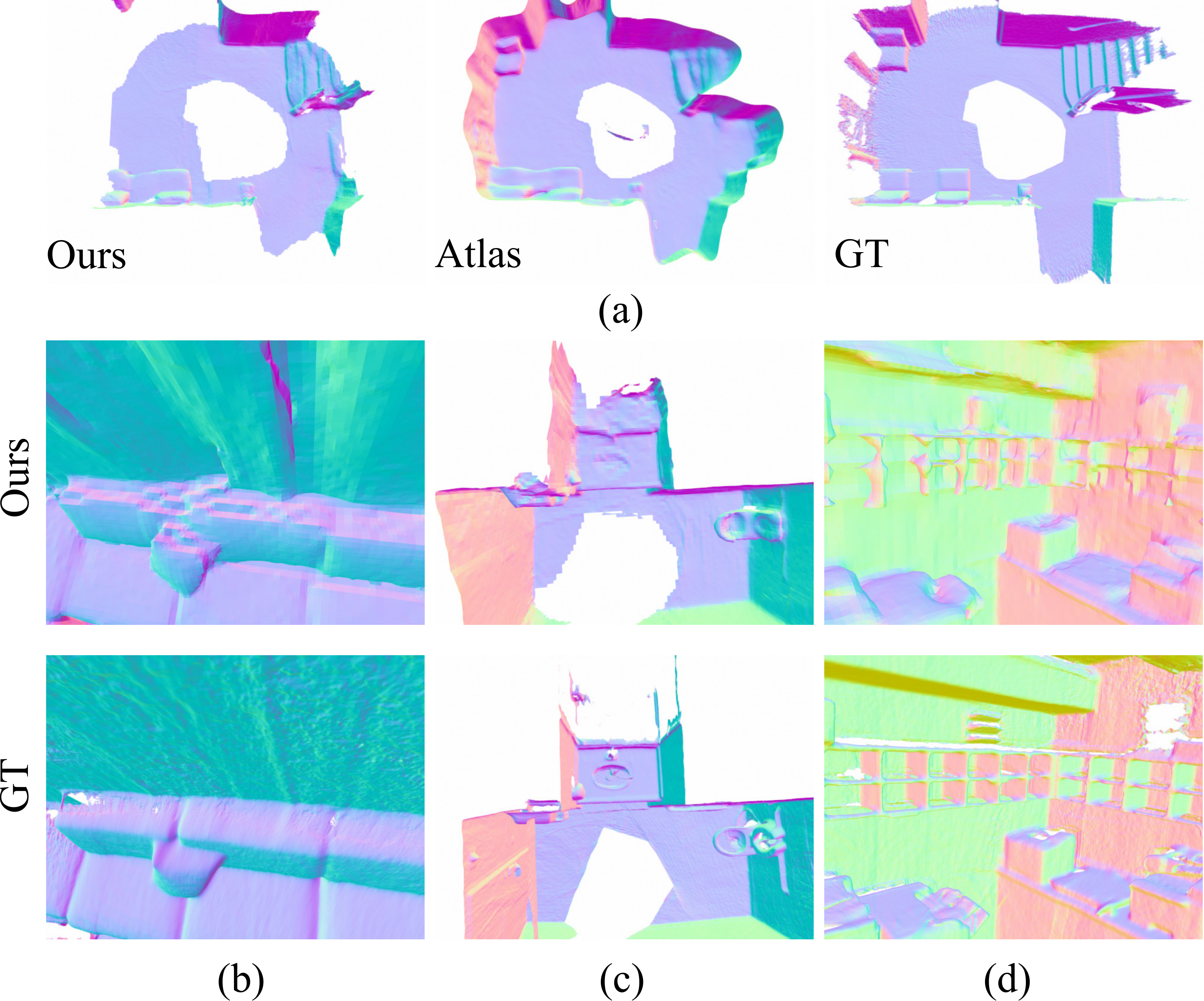}
    \caption{\textbf{Limitations.} \hb{Our method} 
    would \hb{miss slightly more} 
    regions compared with offline method\hb{s} (e.g., Atlas) (a)\hb{. Our method suffers from} 
    inconsistent normal prediction used in optimization (b), mirror reconstruction (c), and complex thin parts (d).}
    \label{fig:failure_cases}
\end{figure}

\subsection{Limitations and Discussion}
Although we formulate the \hb{NISR}  
as the sparse feature volume for \hb{a} more flexible and effective geometry representation, this representation is heavily dependent on the coverage of predicted occupied voxels. Thus, our approach is limited in its ability to complete missing region\hb{s} if some voxels cannot 
be predicted successfully. This missing region is hard to complete during online optimization, leading to a lower recall than the offline methods (see Fig.~\ref{fig:failure_cases}(a)).
Second, the quality of normal prediction influences the online optimization and some inconsistent normal prediction would worsen the reconstruction results (see Fig.~\ref{fig:failure_cases}(b)). \hb{It might be} 
improved by using multi-view normal consistency check to reduce the impact of 
inconsistent normal\hb{s}, or even treating normal vectors as optimization parameters to participate in the optimization.
Third, although our method could achieve finer details of surface reconstruction,  
there is still a gap between the reconstructed models {and} 
depth scans data for complex thin parts (see Fig.~\ref{fig:failure_cases}(d)) and it is still \hb{challenging} 
to reconstruct mirrors (Fig.~\ref{fig:failure_cases}(c)).
Lastly, there is still 
room to improve the timing of our online optimization. The more iterations it optimizes, the better results it could achieve. Therefore, 
techniques to speed up the rendering optimization~\cite{mueller2022instant} can be further adopted to improve the results.

\section{Conclusion}
In this paper, we \hb{introduced} 
MonoNeuralFusion for online 3D scene reconstruction from monocular videos. We formulate a \emph{geometry prior guided} neural implicit scene representation (NISR) with volume rendering to achieving better fine-grained surface reconstruction. We pre-train our NISR with the guidance of geometry priors, 
leading to more effective feature latent vector extraction for 
fine-grained surface reconstruction. To efficiently and effectively render color and normal map\hb{s} in the sparse feature volume, we propose a hierarchical sampling \hb{strategy}, which ensures sampling inside sparse voxels.
Base on these aforementioned, we run our 
online system to incrementally build surface reconstruction, meanwhile, performing the online volume rendering optimization to leverage the geometry cues from normal prediction to enhance geometric details of reconstruction.
We demonstrate that our approach can achieve state-of-the-art quality of indoor scene reconstruction with fine geometric details on different datasets, even using pre-trained weights from the Scannet dataset without further fine-tuning.





\ifCLASSOPTIONcaptionsoff
  \newpage
\fi

\bibliographystyle{IEEEtran}
\bibliography{ref}

\end{document}